\documentclass[letterpaper, 10 pt, journal, twoside]{ieeetran}
\usepackage{graphicx}
\usepackage{times}
\usepackage{amsmath}
\usepackage{amssymb}
\usepackage[utf8]{inputenc}
\usepackage{rgap_notation}
\usepackage{rgap_header}

\newcommand*{\field}[1]{\mathbb{\MakeUppercase{#1}}}		
\newcommand*{\set}[1]{{\mathcal{\MakeUppercase{#1}}}}			

\newcommand*{\sequence}[1]{\mathrm{\MakeUppercase{#1}}}

\newcommand*{\R}{\field{R}} 

\renewcommand{\vec}[1]{{\boldsymbol{\mathbf{#1}}}}
\newcommand*{\mat}[1]{\vec{\MakeUppercase{#1}}}
\newcommand*{\eye}{\mat{I}}							
\newcommand*{\transpose}{\mathsf{T}}

\newcommand*{\observation}{y} 					
\newcommand*{\observations}{\vec{\observation}}
\newcommand*{\obsNoise}{\nu}

\newcommand*{\gp}{\mathcal{GP}}
\newcommand*{\gpMean}{\mu}
\newcommand*{\gpMeanFunction}{m}

\newcommand*{\parameters}{\vec{x}} 
\newcommand*{\ParamSpace}{\set{X}} 

\newcommand*{\state}{s}
\newcommand*{\StateSpace}{\set{S}}
\newcommand*{\action}{a}

\newcommand*{\cost}{c}
\newcommand*{\Cost}{C}
\newcommand*{\trajectory}{\sequence{\state}}
\newcommand*{\stimulus}{v}
\newcommand*{\stimuli}{\sequence{\stimulus}}
\newcommand*{\iReward}{r}       

\newcommand*{\expectation}{\mathbb{E}}

\newcommand*{\normal}{\mathcal{N}}					

\newcommand*{\af}{h} 						
\newcommand*{\Sspace}{\set{S}} 				

\newcommand*{\iterIdx}{t}

\newcommand*{\nIterations}{N}
\newcommand*{\nObs}{n}

\newcommand*{\nTrajectories}{M}
\newcommand*{\nFeatures}{m}

\newcommand*{\anyfunction}{g}

\newcommand*{\iid}{i.i.d.\xspace}

\title{
Heteroscedastic Bayesian Optimisation for Stochastic Model Predictive Control
}

\author{Rel Guzman, Rafael Oliveira, and Fabio Ramos
\thanks{Manuscript received: May, 5, 2020; Revised July, 6, 2020;
Accepted September, 16, 2020.}
\thanks{This paper was recommended for publication by Editor Dongheui Lee upon evaluation of the Associate Editor and Reviewers' comments.
}
\thanks{The authors are with the School of Computer Science, the University of Sydney, Australia. Rafael Oliveira is also with the Australian Research Council Centre for Data Analytics for Resources and Environments (DARE), and Fabio Ramos is also with NVIDIA, USA. { \tt \{rel.guzmanapaza, rafael.oliveira, fabio.ramos\}@sydney.edu.au}\textbf{•}
}
\thanks{Digital Object Identifier (DOI): see top of this page.} 
} 

\begin{document}
\maketitle


\begin{abstract}
Model predictive control (MPC) has been successful in applications involving the control of complex physical systems. This class of controllers leverages the information provided by an approximate model of the system's dynamics to simulate the effect of control actions. MPC methods also present a few hyper-parameters which may require a relatively expensive tuning process by demanding interactions with the physical system. Therefore, we investigate fine-tuning MPC methods in the context of stochastic MPC, which presents extra challenges due to the randomness of the controller's actions. In these scenarios, performance outcomes present noise, which is not homogeneous across the domain of possible hyper-parameter settings, but which varies in an input-dependent way. To address these issues, we propose a Bayesian optimisation framework that accounts for heteroscedastic noise to tune hyper-parameters in control problems. Empirical results on benchmark continuous control tasks and a physical robot support the proposed framework's suitability relative to baselines, which do not take heteroscedasticity into account.
\end{abstract}

\begin{IEEEkeywords} 
Reinforcement Learning; Probability and Statistical Methods; Optimization and Optimal Control
\end{IEEEkeywords} 

\section{Introduction}

\IEEEPARstart{M}{odel} predictive control (MPC) has been a successful approach to optimal control problems in robotics \cite{Howard2010, Paden2016, Williams2017}. Its success relies on incorporating prior information about the system's dynamics into the control loop so that the algorithm may select actions that lead to a predicted optimal performance \cite{Rawlings2017}. However, predictive models are simply numerical approximations to the system's real dynamics, which often render predictions only locally accurate. When combined with non-modelled disturbances, the model's limitations end up compromising predictions over long time horizons. A successful approach to make MPC robust has then been stochastic MPC \cite{Rawlings2017}, such as model predictive path integral (MPPI) controllers \cite{Williams2017journal}.

Stochastic model predictive controllers overcome approximation errors by selecting sequences of actions, which are optimal under random perturbations. To solve optimisation problems, a common approach in stochastic MPC is to roll out multiple trajectories and choose the actions that result in minimum expected cost. For instance, in the case of MPPI, this computation is based on injecting noise into the actions and performing a weighted average over the roll-outs, trying to balance an exploration-exploitation trade-off \cite{Williams2017journal}. However, these steps depend on hyper-parameters, which are hard to tune, as it involves costly interactions with the target system and responses that vary in behaviour. 

Bayesian optimisation (BO) provides an efficient approach to learn hyper-parameters dependent on costly interactions, with applications ranging from robotics to medicine \cite{Shahriari2016}. In applications to control, BO has led to data-efficient frameworks to optimise control policies \cite{Kuindersma2012, Wilson2014}, finding optimal solutions in just a few trials. Despite its success, a major drawback of classical BO is to assume observation noise to be independent and identically distributed (\iid).

\begin{figure}
    \centering
    \includegraphics[width=0.65\columnwidth]{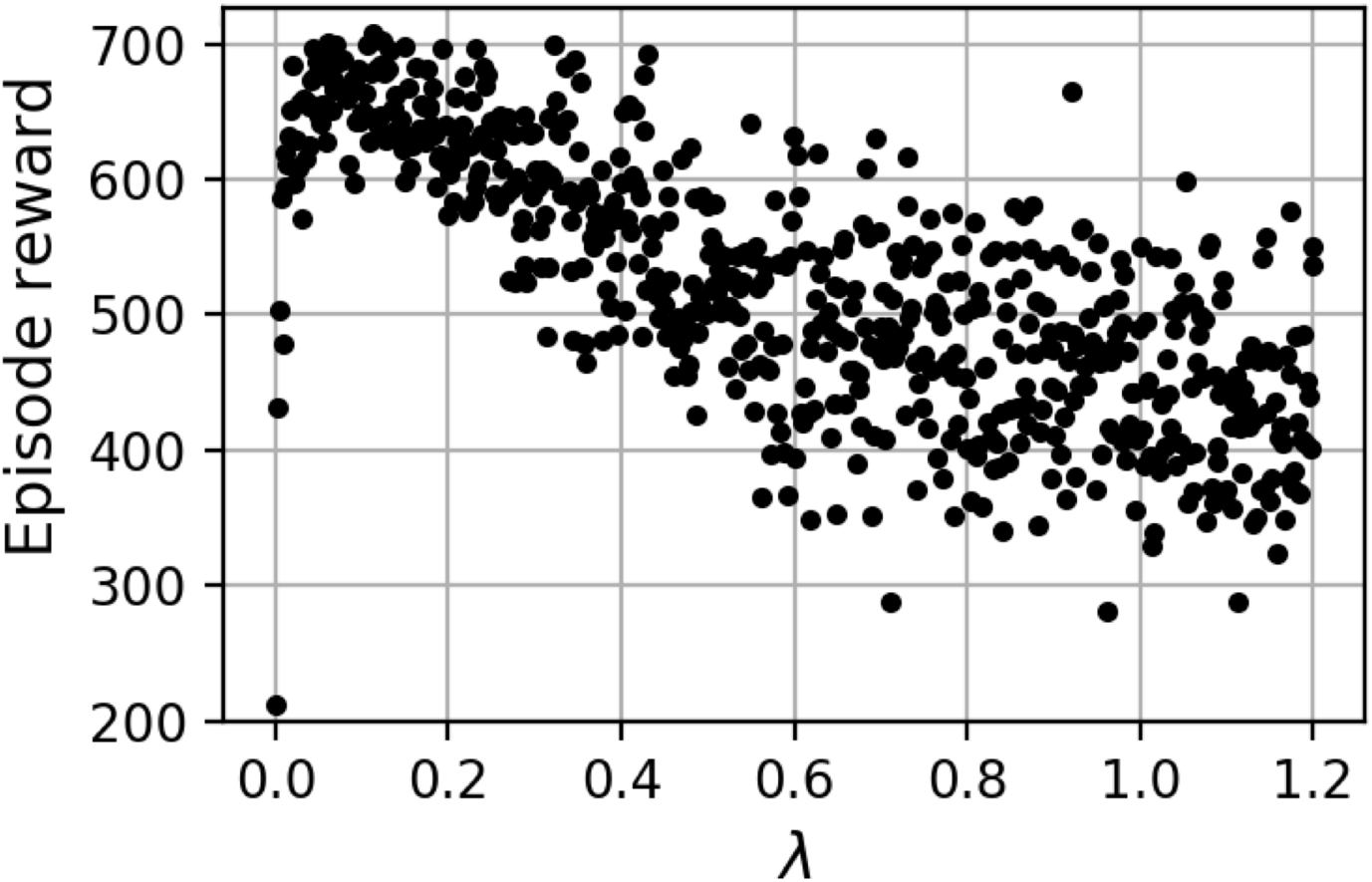}
    \caption{Heteroscedasticity in episode rewards for MPPI in the acrobot task across a range of MPPI temperature $\lambda$ settings.}
    \label{fig:acrobot}
\end{figure}

\autoref{fig:acrobot} presents the performance of MPPI on a classical control problem, the \emph{acrobot} swing-up task \cite{Spong1995}, as a function of the controller's temperature hyper-parameter. As the plot shows, the rewards' variance across the range of temperature values is not uniform. Noise in the highest rewards region is significantly less than elsewhere, evidencing an input-dependent behaviour known as \emph{heteroscedasticity}. This behaviour has been approached in different ways in the BO literature \cite{Kuindersma2012, Wilson2017}, leading to performance improvements.

This paper investigates the effect of heteroscedasticity in the tuning of hyper-parameters for optimal controllers via BO. In particular, we analyse MPPI's performance as a function of its hyper-parameters and propose methods to account for heteroscedastic noise. We make the following contributions:
\begin{itemize}
\item a framework to tune stochastic MPC via heteroscedastic Bayesian optimisation;
\item a class of parametric models for heteroscedastic noise in the controller's response distribution; and
\item experimental results on a range of benchmark continuous control problems in simulated and real scenarios.
\end{itemize}

The next sections start by discussing related work in control, reinforcement learning, and Bayesian optimisation. We follow with background on MPPI and BO. \autoref{sec:method} then describes our proposed methodology. In \autoref{sec:experiments}, we present experimental results, and \autoref{sec:conclusion} concludes the paper.

\section{Related Work}
\label{sec:related}

To start with, model predictive control (MPC) and model-based reinforcement learning (RL) both approach control problems \cite{Gorges2017}, with a noticeable similarity in the use of an assumed or learnt dynamical model of the system. Learning a controller is often constrained by the number of interactions with the environment due to inherent real-world restrictions, such as energy and mechanical wear \cite{Deisenroth2013}. Model-based approaches bypass most of these limitations by using information from the model \cite{Chua2018, Mahmood2018}.
Conversely, many existing model-free algorithms become impractical in challenging real-world scenarios, such as autonomous vehicles \cite{Polydoros2017}.

Traditional model predictive control methods usually become inefficient when dealing with highly non-linear dynamics and non-convex reward functions \cite{Williams2017journal, Wang2019}. Some state-of-the-art approaches can efficiently adapt to such challenging stochastic environments with sampling-based methods. For example, a flexible data-driven sampling-based MPC method is Model Predictive Path Integral (MPPI) \cite{Williams2016}. 

MPPI is a type of optimal controller that selects controls via an information-theoretic sampling-based algorithm \cite{Williams2016,Williams2017}. Like any other optimisation algorithm, however, MPPI has hyper-parameters that balance exploration and exploitation, which raises the question of how to tune them. Hyper-parameters often have to be optimised according to the task and learning behaviours, leading to settings that are not necessarily transferable across tasks \cite{Mahmood2018}. The work in \cite{Liang2019} proposes online hyper-parameter optimisation to improve MPPI's performance. Their method consists of a meta-learning approach to learn the dynamics offline and adjust to disturbances online with an adaptive temperature coefficient.

Bayesian optimisation has been widely applied to hyper-parameter tuning \cite{Snoek2012, Shahriari2016}. BO performs inference about a latent objective function by modelling it as a Gaussian process (GP) \cite{Rasmussen2006}. For the GP, it is commonly assumed that observation noise is \iid Gaussian across the search space. 

Heteroscedastic noise with parametric noise models can be learnt via maximum likelihood \cite{Wilson2017}. A Bayesian approach is to add a second GP prior to the log-variance of the noise model \cite{Goldberg1998}. The resulting stochastic process is no longer Gaussian and requires Markov chain Monte Carlo \cite{Andrieu2003} for inference. Approximate inference methods have also been proposed to reduce the computational overhead by using variational inference \cite{Kersting2007, Lazaro-Gredilla2011}. Unlike the computationally expensive variational approximation from \cite{Kuindersma2012}, we use a parametric formulation with heteroscedastic noise learnt by maximising the GP marginal likelihood.

The use of flexible non-parametric priors, such as GPs for the noise model, leads to an increase in computational complexity and a resulting model which is not exactly a GP, but only approximated as such. In this paper, we take a simpler approach, using a flexible parametric noise model to encode prior knowledge about the noise process in applications of stochastic MPC.

\section{Background}
\label{sec:background}

To facilitate our discussion, we first introduce background on model predictive control, Gaussian processes, and Bayesian optimisation, alongside their respective notation.

\subsection{Transition Model and Model-based Control}

We consider a dynamical system with states $\state \in \StateSpace$ and admissible controls (or actions) $a \in \mathcal{A}$ where the state follows Markovian dynamics, $s_{t+1}=f\left(s_{t}, a_{t}\right)$, with a transition function $f$ and a reward function $\iReward$ that measures how well the system is doing given a state and action $\iReward : \mathcal{S} \times \mathcal{A} \to \R$.
Although the true dynamics are usually unknown, a model of the transition function can be learned or assumed from expert knowledge. 

\subsection{Stochastic Model Predictive Control}

\begin{figure}[t]
\centering
\tikzset{every picture/.style={line width=0.75pt}} 
\begin{tikzpicture}[x=0.75pt,y=0.75pt,yscale=-0.65,xscale=0.8]
\draw  [color={rgb, 255:red, 0; green, 0; blue, 0 }  ,draw opacity=1 ][fill={rgb, 255:red, 233; green, 233; blue, 233 }  ,fill opacity=1 ][line width=1.5]  (9.38,36.09) .. controls (9.38,20.99) and (21.62,8.75) .. (36.72,8.75) -- (230.04,8.75) .. controls (245.14,8.75) and (257.38,20.99) .. (257.38,36.09) -- (257.38,118.11) .. controls (257.38,133.21) and (245.14,145.45) .. (230.04,145.45) -- (36.72,145.45) .. controls (21.62,145.45) and (9.38,133.21) .. (9.38,118.11) -- cycle ;
\draw  [fill={rgb, 255:red, 255; green, 255; blue, 255 }  ,fill opacity=1 ] (69.79,98.88) .. controls (69.79,93.36) and (74.26,88.89) .. (79.78,88.89) -- (184.8,88.89) .. controls (190.32,88.89) and (194.79,93.36) .. (194.79,98.88) -- (194.79,128.85) .. controls (194.79,134.37) and (190.32,138.84) .. (184.8,138.84) -- (79.78,138.84) .. controls (74.26,138.84) and (69.79,134.37) .. (69.79,128.85) -- cycle ;
\draw  [fill={rgb, 255:red, 255; green, 255; blue, 255 }  ,fill opacity=1 ] (70.03,46.68) .. controls (70.03,41.97) and (73.85,38.14) .. (78.56,38.14) -- (186.26,38.14) .. controls (190.97,38.14) and (194.79,41.97) .. (194.79,46.68) -- (194.79,72.28) .. controls (194.79,76.99) and (190.97,80.81) .. (186.26,80.81) -- (78.56,80.81) .. controls (73.85,80.81) and (70.03,76.99) .. (70.03,72.28) -- cycle ;
\draw    (69.17,113.19) .. controls (42.92,94.66) and (41.4,79.24) .. (67.01,59.43) ;
\draw [shift={(69.03,57.89)}, rotate = 503.41] [fill={rgb, 255:red, 0; green, 0; blue, 0 }  ][line width=0.08]  [draw opacity=0] (8.93,-4.29) -- (0,0) -- (8.93,4.29) -- cycle    ;
\draw    (195.03,58.89) .. controls (221.62,80.22) and (220.66,92.43) .. (197.39,111.41) ;
\draw [shift={(195.17,113.19)}, rotate = 321.78] [fill={rgb, 255:red, 0; green, 0; blue, 0 }  ][line width=0.08]  [draw opacity=0] (8.93,-4.29) -- (0,0) -- (8.93,4.29) -- cycle    ;
\draw  [color={rgb, 255:red, 0; green, 0; blue, 0 }  ,draw opacity=1 ][line width=1.5]  (318.4,55.38) .. controls (318.4,47.98) and (324.4,41.98) .. (331.8,41.98) -- (400,41.98) .. controls (407.4,41.98) and (413.4,47.98) .. (413.4,55.38) -- (413.4,95.58) .. controls (413.4,102.98) and (407.4,108.98) .. (400,108.98) -- (331.8,108.98) .. controls (324.4,108.98) and (318.4,102.98) .. (318.4,95.58) -- cycle ;
\draw [line width=1.5]    (257.34,55.38) -- (314.4,55.38) ;
\draw [shift={(318.4,55.38)}, rotate = 180] [fill={rgb, 255:red, 0; green, 0; blue, 0 }  ][line width=0.08]  [draw opacity=0] (13.4,-6.43) -- (0,0) -- (13.4,6.44) -- (8.9,0) -- cycle    ;
\draw [line width=1.5]    (261.54,98.38) -- (318.4,98.38) ;
\draw [shift={(257.54,98.38)}, rotate = 0] [fill={rgb, 255:red, 0; green, 0; blue, 0 }  ][line width=0.08]  [draw opacity=0] (13.4,-6.43) -- (0,0) -- (13.4,6.44) -- (8.9,0) -- cycle    ;

\draw (133,120.68) node  [font=\normalsize]  {$s_{t+1} =f( s_{t} ,a_{t})$};
\draw (132.41,48.99) node  [font=\footnotesize] [align=left] {\textbf{{\small Optimisation}}};
\draw (132.29,99.61) node  [font=\footnotesize] [align=left] {\textbf{{\small Transition Model}}};
\draw (234,85) node  [font=\footnotesize] [align=left] {Action};
\draw (32,85) node  [font=\footnotesize] [align=left] {State};
\draw (137,23.42) node   [align=left] {\textbf{{\small Model Predictive Control}}};
\draw (365.9,75.48) node   [align=left] {\textbf{{\small System}}};
\draw (290,43.16) node  [font=\small]  {$a^{*}_{t}$};
\draw (293,115.96) node  [font=\small]  {$s^{( s)}_{t+1}$};
\draw (84,57) node [anchor=north west][inner sep=0.75pt]  [font=\footnotesize] [align=left] {Reward Function};

\end{tikzpicture}
\caption{Model predictive control loop. MPC optimises next actions according to a reward function and a transition model $f$ within a time horizon. Then, the next action $a^{*}_t$ is received by the actuator and the system moves to a new state $s_{t+1}$.}
\label{fig:mpc}
\end{figure}
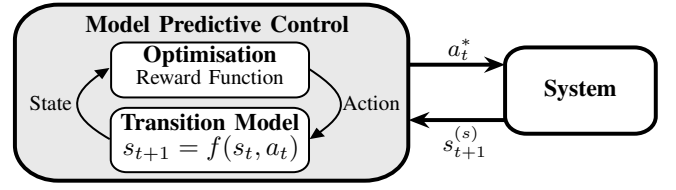

Model predictive control, also known as receding horizon control, is a class of algorithms that operate by optimising sequences of actions over approximate models of a system. MPC solves an optimisation problem up to a horizon $T$ constrained by a dynamical system $f$ and then executes the next best action. The diagram in \autoref{fig:mpc} describes the interaction between the system and the controller in MPC.

A flexible MPC method that optimises controls as an information-theoretic sampling-based algorithm is model predictive path integral (MPPI) \cite{Williams2016}. At time step $t$, MPPI outputs sequences of noise-perturbed controls $\stimuli_t = \{\stimulus_{i}\}_{i=t}^{t+T}$, where $\stimulus_i = \action_i^* + \epsilon_i$ and $\epsilon_i \sim \normal(0, \sigma_\epsilon^2)$, based on a rollover sequence of optimal actions $\{\action_i^*\}_{i=t}^{t+T}$ that start as 0.

When applied to a model of the system, each control sequence results in a sequence of states $\trajectory_t = \{\state_{t+i}\}_{i=1}^T$ with cost determined by a function associated with the control task:
\begin{align}
\Cost(\trajectory_t)=\phi\left(s_{t+T}\right) + \sum_{i=1}^{T-1} \cost(s_{t+i})~,
\end{align}
where $\cost:\StateSpace\to\R^+$ is an instant cost function, and $\phi:\StateSpace\to\R^+$ represents terminal cost.
Based on $\nTrajectories$ rollouts, MPPI updates the sequence of optimal actions and weights \cite{Williams2017}:
\begin{equation}
\action_{i}^* \leftarrow \action_i^* + \sum_{j=1}^{\nTrajectories} w(\stimuli_t^{j})\epsilon_i^{j}\,, \quad j \in \{1,\ldots, \nTrajectories\}\,,
\end{equation}
\begin{equation}
\label{eq:weights}
w(\stimuli_t) = \frac{1}{\eta} \exp\left(-\frac{1}{\lambda}\left(\Cost(\trajectory_t) + \frac{\lambda}{\sigma_\epsilon^2} \sum_{i=t}^{t+T}\action_{i}^*\cdot\stimulus_{i}\right)\right)\,,
\end{equation}
and $\eta$ is a normalisation constant, so that $\sum_{j=1}^\nTrajectories w(\stimuli_t^j) = 1$.
MPPI then applies the first action in the sequence to the real system, discards it and appends a new random action to the end of the sequence. This process repeats every time step.

The second hyper-parameter appears in \eqref{eq:weights} and is called temperature $\lambda \in \R^+$. Intuitively, a higher variance $\sigma_\epsilon^2$ results in more varying and forceful actions, while  $\lambda \to 0$ leads the optimal distribution to place all its mass on a single trajectory. Conversely, $\lambda \to \infty$ would make all trajectories have similar probabilities of occurrence~\cite{Williams2018a}.

Both hyper-parameters control exploration and exploitation. Higher $\lambda$ or $\sigma_\epsilon$ result in more exploration in the action space, while lower $\lambda$ or $\sigma_\epsilon$ result in more exploitation. That raises the question of how to find the best hyper-parameter settings, which may have to be tuned according to the task.

\subsection{Gaussian Processes}
\label{sec:gp}
A Gaussian process \cite{Rasmussen2006} represents a probability distribution over a space of functions. A GP prior over a function $\anyfunction:\ParamSpace\to\R$ is completely specified by a mean $\gpMeanFunction:\ParamSpace\to\R$ and a positive-definite covariance function, $k:\ParamSpace\times\ParamSpace\to\R$. Under the GP prior, the values of $\anyfunction$ at a finite collection of points $\{\parameters_i\}_{i=1}^\nObs \subset\ParamSpace$ follow a multivariate normal distribution $\anyfunction(\mat\parameters) \sim \normal(\vec{\gpMeanFunction},\mat K)$, where $\anyfunction(\mat\parameters) = [\anyfunction(\parameters_1),\dots,\anyfunction(\parameters_\nObs)]^\transpose$, $\vec{\gpMeanFunction} := \gpMeanFunction(\mat\parameters)$, and $\mat K$ is the $\nObs$-by-$\nObs$ covariance matrix given by $[\mat K]_{i,j} = k(\parameters_i,\parameters_j)$.

\subsubsection{Inference}
Now suppose we observe $\observations\in\R^\nObs$, where each $\observation_i = \anyfunction(\parameters_i) + \obsNoise_i$ represents a function evaluation corrupted by jointly Gaussian noise $\vec\obsNoise \sim \normal(\vec{0}, \mat\Sigma_\obsNoise)$. The joint distribution of the observations and the function value at a point $\parameters\in\ParamSpace$ is then given by:
\begin{equation}
\left[
\begin{array}{c}
\observations\\
\anyfunction(\parameters)
\end{array}
\right]
\sim
\normal
\left(
\left[
\begin{array}{c}
     \vec{\gpMeanFunction}\\
     \gpMeanFunction(\parameters)
\end{array}
\right]
\,,
\left[
\begin{array}{cc}
\mat K + \mat\Sigma_\obsNoise & \vec{k}(\parameters)\\
\vec{k}(\parameters)^\transpose & k(\parameters,\parameters)
\end{array}
\right]
\right)\,,
\end{equation}
where $\vec{k}(\parameters) := [k(\parameters, \parameters_1), \dots, k(\parameters, \parameters_\nObs)]^\transpose$,
Conditioning $\anyfunction(\parameters)$ on the observations yields a Gaussian predictive distribution $\anyfunction(\parameters)|\observations \sim \normal(\gpMean(\parameters),\sigma^2(\parameters))$,
where:
\begin{align}
\gpMean(\parameters) &= \gpMeanFunction(\parameters) + \vec{k}(\parameters)^\transpose(\mat K + \mat\Sigma_\obsNoise)^{-1}(\observations-\vec{\gpMeanFunction})\\
\sigma^2(\parameters) &= k(\parameters, \parameters) - \vec{k}(\parameters)^\transpose(\mat K + \mat\Sigma_\obsNoise)^{-1}\vec{k}(\parameters)~,
\label{eq:posteriors}
\end{align}
allowing us to infer function values at unobserved locations.

\subsubsection{Noise model}
In general, observation noise $\obsNoise$ is assumed to be \emph{homoscedastic}, which means its distribution is not dependent on the inputs $\parameters$. However, many applications present noise with a \emph{heteroscedastic} behaviour, i.e. the noise distribution varies across the domain $\ParamSpace$. Under the Gaussian assumption, observation noise is simply another (zero-mean) Gaussian process with covariance function $k_\obsNoise:\ParamSpace\times\ParamSpace\to\R$, so that $[\mat\Sigma_\obsNoise]_{i,j} := k_\obsNoise(\parameters_i,\parameters_j)$. In the \emph{homoscedastic} case, the noise covariance function is simply $k_\obsNoise(\parameters, \parameters) := \sigma^2_\obsNoise$, where $\sigma_\obsNoise\in\R$ is constant, and $k_\obsNoise(\parameters, \parameters') = 0$ for $\parameters\neq\parameters'$, yielding the classic $\mat\Sigma_\obsNoise = \sigma^2_\obsNoise\eye$. More generally, however, $k_\obsNoise$ can be an arbitrary positive-definite covariance function.

\subsection{Bayesian Optimisation}
\label{sec:bo}
Consider the problem of searching for the global optimum of a function $\anyfunction:\ParamSpace \to \mathbb{R}$ over a given compact search space $\Sspace \subset \ParamSpace$ such as determining
$\parameters^* \in \argmax_{\parameters \in \Sspace} \anyfunction(\parameters)$. Assume that $\anyfunction$ is possibly non-convex and only partially observable via noisy estimates $\observation_\iterIdx = \anyfunction(\parameters_\iterIdx) + \obsNoise_\iterIdx$ with $\obsNoise_\iterIdx \sim \normal(0,\sigma_{\obsNoise_\iterIdx}^2)$. In addition, we can only observe the function up to $\nIterations$ times.

Bayesian optimisation \cite{Shahriari2016} assumes that $\anyfunction$ is a random variable itself and models it as a stochastic process, which is usually a GP, indexed by $\ParamSpace$. To select points at which to observe $\anyfunction$, BO uses an acquisition function $\af(\parameters)$ as a guide that incorporates prior information provided by the GP model and the observations. Each query point $\parameters_t \in \Sspace$ is then selected by maximising $\af$. After collecting an observation $\observation_t$, BO updates the GP model with the pair $(\parameters_t,\observation_t)$ and starts the next iteration with an improved belief about $f$. The BO loop repeats until we reach the given budget of $\nIterations$ evaluations of the objective function. See \autoref{alg:bayesopt} for a summary.

\begin{algorithm}[t]
    \caption{Bayesian Optimisation}
    \label{alg:bayesopt}
    \textbf{Input:} Sampling iterations $n$; search space $\Sspace$\\
    \textbf{Output:} $\left(\mathbf{x}^{*}, y^{*}\right)$\\
    \For{$t = 1$ \KwTo $n$}{
        Fit a GP model $\mathcal{M}$ on the data $\mathcal{D}_{1:t}$\\
        Find $\mathbf{x}_{t}=\argmax_{\parameters \in \Sspace} \af(\mathbf{x},  \mathcal{M}, \mathcal{D}_{1:t})$\\
        $y_t \leftarrow$ Evaluate the objective function at $\mathbf{x}_t$\\
        $\mathcal{D}_{1:t+1} = \mathcal{D}_{1:t} \cup \{(\mathbf{x}_t, y_t)\}$
}
\end{algorithm}
 
The acquisition function $\af$ determines which values to sample next. A common and simple acquisition function is the upper confidence bound (UCB) \cite{Cox1992}: 
\begin{equation}
\label{eq:ucb}
\af_\text{ucb}(\vec{x}):=\mu(\vec{x}) + \kappa \sigma(\vec{x})~,
\end{equation}
where $\kappa \in \R^+$ is a balance factor. UCB allows balancing exploration and exploitation by valuing points where there is high uncertainty (exploration) or where the GP predictive mean is high (exploitation). Keeping the balance factor $\kappa$ biased towards exploration avoids local minima.

\section{METHODOLOGY}
\label{sec:method}
Hyper-parameter optimisation is often done manually by following prior experience from similar problems. Here we deal with automatically finding MPC hyper-parameters that could lead to significant differences in performance.

An MPC controller can be treated as a black-box model that receives hyper-parameters $\parameters$, a model of the transition dynamics $f$, a time horizon $T$, and the number of trajectory roll-outs $M$. In this paper, we are not concerned with tuning $M$ and $T$, which mostly depend on computational resources. We instead focus on tuning $\vec{x}$ via heteroscedastic BO.

\begin{figure}
\centering
\tikzset{every picture/.style={line width=0.75pt}} 
\begin{tikzpicture}[x=0.75pt,y=0.75pt,yscale=-0.65,xscale=0.95]

\draw  [color={rgb, 255:red, 0; green, 0; blue, 0 }  ,draw opacity=1 ][line width=1.5]  (37.97,511.07) .. controls (37.97,501.76) and (45.52,494.21) .. (54.83,494.21) -- (281.11,494.21) .. controls (290.42,494.21) and (297.97,501.76) .. (297.97,511.07) -- (297.97,561.65) .. controls (297.97,570.96) and (290.42,578.51) .. (281.11,578.51) -- (54.83,578.51) .. controls (45.52,578.51) and (37.97,570.96) .. (37.97,561.65) -- cycle ;
\draw  [fill={rgb, 255:red, 233; green, 233; blue, 233 }  ,fill opacity=1 ][line width=1.5]  (1.76,394.55) .. controls (1.76,384.63) and (9.8,376.59) .. (19.72,376.59) -- (304.8,376.59) .. controls (314.72,376.59) and (322.76,384.63) .. (322.76,394.55) -- (322.76,448.43) .. controls (322.76,458.35) and (314.72,466.39) .. (304.8,466.39) -- (19.72,466.39) .. controls (9.8,466.39) and (1.76,458.35) .. (1.76,448.43) -- cycle ;
\draw  [color={rgb, 255:red, 0; green, 0; blue, 0 }  ,draw opacity=1 ][fill={rgb, 255:red, 255; green, 255; blue, 255 }  ,fill opacity=1 ][line width=1.5]  (169.24,413.83) .. controls (169.24,407.94) and (174.02,403.16) .. (179.91,403.16) -- (304.57,403.16) .. controls (310.47,403.16) and (315.24,407.94) .. (315.24,413.83) -- (315.24,445.84) .. controls (315.24,451.73) and (310.47,456.51) .. (304.57,456.51) -- (179.91,456.51) .. controls (174.02,456.51) and (169.24,451.73) .. (169.24,445.84) -- cycle ;
\draw [line width=1.5]    (116.27,420.38) -- (165.24,420.38) ;
\draw [shift={(169.24,420.38)}, rotate = 180] [fill={rgb, 255:red, 0; green, 0; blue, 0 }  ][line width=0.08]  [draw opacity=0] (13.4,-6.43) -- (0,0) -- (13.4,6.44) -- (8.9,0) -- cycle    ;
\draw [line width=1.5]    (129.65,447.38) -- (169.24,447.38) ;
\draw [shift={(125.65,447.38)}, rotate = 0] [fill={rgb, 255:red, 0; green, 0; blue, 0 }  ][line width=0.08]  [draw opacity=0] (13.4,-6.43) -- (0,0) -- (13.4,6.44) -- (8.9,0) -- cycle    ;
\draw [line width=1.5]    (170.48,583.51) -- (170.48,600.51) -- (308.97,600.51) -- (307.53,466.87) ;
\draw [shift={(170.48,579.51)}, rotate = 90] [fill={rgb, 255:red, 0; green, 0; blue, 0 }  ][line width=0.08]  [draw opacity=0] (13.4,-6.43) -- (0,0) -- (13.4,6.44) -- (8.9,0) -- cycle    ;
\draw [line width=1.5]    (168.88,492.39) -- (168.88,471.34) ;
\draw [shift={(168.88,467.34)}, rotate = 450] [fill={rgb, 255:red, 0; green, 0; blue, 0 }  ][line width=0.08]  [draw opacity=0] (13.4,-6.43) -- (0,0) -- (13.4,6.44) -- (8.9,0) -- cycle    ;
\draw  [fill={rgb, 255:red, 255; green, 255; blue, 255 }  ,fill opacity=1 ][line width=1.5]  (9.97,414.11) .. controls (9.97,408.13) and (14.81,403.29) .. (20.79,403.29) -- (114.83,403.29) .. controls (120.81,403.29) and (125.65,408.13) .. (125.65,414.11) -- (125.65,446.57) .. controls (125.65,452.55) and (120.81,457.39) .. (114.83,457.39) -- (20.79,457.39) .. controls (14.81,457.39) and (9.97,452.55) .. (9.97,446.57) -- cycle ;

\draw (243.16,415.85) node   [align=left] {\textbf{{\small System}}};
\draw (146.43,408.16) node  [font=\small]  {$a^{*}_{i}$};
\draw (151.98,457.96) node  [font=\small]  {$s_{i+1}$};
\draw (167.97,504.42) node   [align=left] {\textbf{{\small Heteroscedastic BO}}};
\draw (188.88,478.68) node  [font=\small]  {$\mathbf{x}^{*} \ $};
\draw (295.26,480.29) node  [font=\small]  {$G_{t\ }$};
\draw (169.88,544) node  [font=\small] [align=left] {$\bullet $ Set predefined noise model $\displaystyle q(\mathbf{x})$\\$\bullet $ Fit GP reward model with $\displaystyle \left(\mathbf{x}_{1:t} ,g_{1:t} ,\mathnormal{q}\right)$\\$\bullet $ Find $\displaystyle \mathbf{x}^{*}$};
\draw (69.05,437.26) node  [font=\small]  {$a^{*}_{i} =\text{MPC} (s_{i} ,\mathbf{x}^{*} ,f)$};
\draw (68.02,415.85) node   [align=left] {\textbf{Controller}};
\draw (243.63,437.26) node  [font=\small]  {$r_{i} =\text{SendToActuators} (a^{*}_{i} )$};
\draw (164.12,389.42) node   [align=left] {\textbf{Control Loop}};

\end{tikzpicture}
\caption{General overview of MPC optimisation with BO.}
\label{fig:block}
\end{figure}
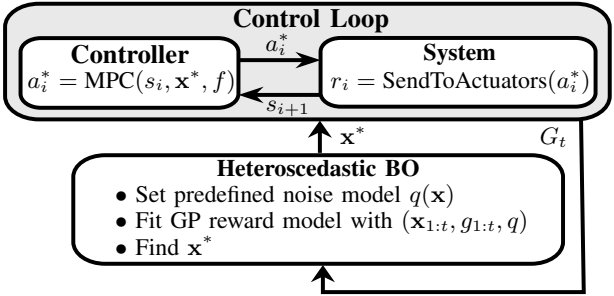

\subsection{Expected Cumulative Reward Function}
As a performance indicator, our framework consists of accumulating instant rewards over episodes, as shown in \autoref{fig:block}. An episode is defined as a sequence of controller-system interactions $\{s_i, a_i, s_{i+1}, a_{i+1}, \ldots\}$, and each action $a_i$ returns a respective reward $r_i$ from the system. We deal with fixed-length episodes, with a control loop executed up to time $n_e$.  At each time step, the MPC's optimal action is sent to the system actuators, obtaining a reward $r_i$ that is accumulated along the episode as $g = \sum_{i}^{n_e} r_i$. However, due to a number of factors, such as an arbitrary initial state, rewards are stochastic.

Our objective is to maximise the expected cumulative reward $\hat g(\parameters) := \expectation\left[g(\parameters)\right]$ of an episode as a function of the MPC controller hyper-parameters by finding:
\begin{equation}
    \parameters^* \in \argmax_{\parameters\in\Sspace} \hat g(\parameters)~.
\end{equation}
Computing the expected cumulative reward of an episode is intractable, as it requires marginalising over many variables, including the stochastic behaviour of the controller itself. In practice, instead, we use observations $y = \frac{1}{n_r} \sum_{j=1}^{n_r} g_j(\parameters)$ based on a finite number of episodes $n_r$. We model the expected cumulative reward as $\hat g \sim \gp(\gpMeanFunction, k)$ with an independent noise process for the observations $\obsNoise \sim \gp(0, k_\obsNoise)$.

\subsection{Noise Model}

A constant noise variance is an unrealistic assumption in many practical applications. In the case of MPPI, \autoref{fig:acrobot} shows episode rewards over a range of settings for the temperature hyper-parameter $\lambda$ in the Acrobot task. The plot shows a clear increase in noise levels when increasing $\lambda$, suggesting heteroscedasticity.

\begin{figure}[t]
\centering
\includegraphics[width=0.49\columnwidth]{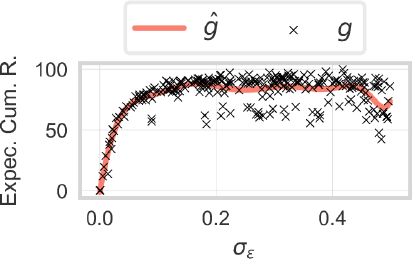}
\includegraphics[width=0.49\columnwidth]{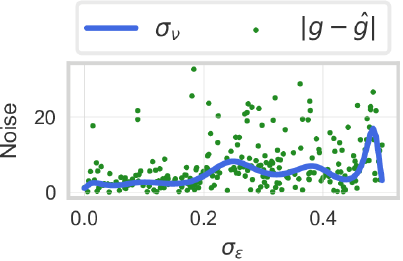}
\caption{Example of a 10-degree polynomial regression model $\hat g$ fitting the expected cumulative reward function in the upper figure, while an estimate for the noise model $\sigma_\nu$ is represented as the blue curve in the lower figure.}
\label{fig:fitted}
\end{figure}

In our context, noise corresponds to the difference between the observation $y$ and the expected cumulative reward for a given setting $\parameters$, i.e.\ $\obsNoise(\parameters) := y - \hat g(\parameters)$. Recalling \autoref{sec:gp}, $\obsNoise(\parameters)$ can be modelled as an independent zero-mean Gaussian process with covariance function $k_\obsNoise$. We consider episodes to be executed independently, so that $k_\obsNoise(\parameters,\parameters') = 0$ for $\parameters \neq \parameters'$. Our concern is then modelling $k_\obsNoise(\parameters, \parameters) = \sigma_\obsNoise^2(\parameters)$.

In this paper, we assume a flexible parametric form for the noise model:
\begin{align}
    \sigma_\obsNoise(\parameters) &= z \cdot \exp\left(\vec{\beta}^\transpose\vec{\phi}(\parameters)\right) + \zeta~,
    \label{eq:noise-model}
\end{align}
where $\vec\beta\in \R^\nFeatures$, $\zeta \geq 0$ and $\vec{\phi}: \ParamSpace\to\R^\nFeatures$ is a feature map. A scalar factor $z$ is added to determine variations of the cumulative reward around its expected value. Small values of $z$ produce expected cumulative reward functions that are close to their mean, and larger values allow more variation. If $z$ is large, the modelled expected cumulative reward function will be able to account for more outliers. A smooth feature map $\phi$ allows the noise model to fit the gradually changing noise variance, as in \autoref{fig:fitted}, with no sharp changes across the search space, and the exponential function ensures the positive-definiteness of $k_\obsNoise$.

The offset term $\zeta$ accounts for any homoscedastic component in the noise process, representing a minimum amount of noise to expect. The exponential term includes a generalised linear model $\vec{\beta}^\transpose\vec{\phi}(\parameters)$ which allows us to vary the noise distribution as a function of the input. The choice of feature map $\vec{\phi}$ is arbitrary. For instance, polynomial features allow us to capture general non-linear trends, while kernel-based features allow us to model localised behaviour.

Having a parametric form for $k_\obsNoise$, there are multiple methods to learn a suitable noise model from data. In this paper, we deal with two of them. One can either directly maximise the log-marginal likelihood \cite{Rasmussen2006} of the GP representing $\hat g \sim \gp(m, k)$ alongside other GP hyper-parameters $\theta$ or learn noise parameters separately in a two-stage regression problem, which is described further below.

From a set of randomly sampled inputs $\parameters_i \in \Sspace$, we can approximate the expected reward $\hat g$ with the flexible generalised linear regression model based on the feature map $\vec{\phi}: \ParamSpace \to \R^\nFeatures$ and learnt weights $\vec\alpha \in\R^\nFeatures$ as: 
\begin{equation}
\begin{split}
\hat{g}(\parameters) \approx \vec{\alpha}^\transpose \vec{\phi}(\parameters)~.
\end{split}
\label{eq:reward-approx}
\end{equation}
With this estimate $\hat g$ we then fit the residuals  $q(\parameters) := |g(\parameters) - \hat{g}(\parameters)|$ with \eqref{eq:noise-model} as a regression problem.

\subsection{Bayesian Controller Optimisation}

We optimise the controller with BO, a global optimisation method, by maximising the episodic or cumulative reward $g$ dependent on controller hyper-parameters $\parameters$ to solve $\parameters^* = \argmax g(\parameters)$. Considering $g$ is stochastic, we maximise the expected cumulative reward $\hat g_t= \expectation\left[g(\parameters_t)\right]$.

\begin{algorithm}[t]
    \caption{Bayesian Controller Optimisation}
    \label{alg:boGlob}
    \textbf{Input:} Controller hyper-parameter search space $\Sspace$,\\
    GP hyper-parameters $\theta$,\\
    number of BO iterations $n_{BO}$,\\
    number of time-steps in an episode $n_e$\\
    \textbf{Output:} $\left(\mathbf{x}^{*}, \hat g^{*}\right)$\\     
    \For{$t = 1$ \KwTo $n_{BO}$}{
        Fit GP model $\mathcal{M}$ with $\mathcal{D}_{1:t}$\\
        Find $\mathbf{x}_{t}=\argmax_{\parameters \in \Sspace} \af(\mathbf{x},  \mathcal{M}, \mathcal{D}_{1:t})$\\
        \For{$j = 1$ \KwTo $n_r$}{
            $g_j(\vec x_t) = 0$\\
            \For{$i = 1$ \KwTo $n_e$}{
                $a_{i}^* = \text{MPC}(\mathbf{x}_{t}, f)$\\
                $r_{i} = \text{SendToActuators}(a_{i}^*)$\\
                $g_j(\vec x_t) \mathrel{+}= r_{i}$
            }
        }
        $y_t = 1/n_r \sum_j\left[g_j(\vec x_t)\right]$\\
        $\mathcal{D}_{1:t+1} = \mathcal{D}_{1:t} \cup \{(\mathbf{x}_t, y_{t})\}$
    }
\end{algorithm}

The controller hyper-parameters are optimised following \autoref{alg:boGlob}.
At each BO iteration, we fit the GP model $\mathcal{M}$ with observations collected up to the current iteration $t$. Next, we select controller hyper-parameters $\vec{x}_{t}$ by maximising the acquisition function $h$ with a global optimisation method.
We then compute the expected cumulative reward $\hat g_t$ empirically by averaging the cumulative rewards obtained after $n_r$ episodes of $n_e$ time steps each.
At each time-step $i$, an optimal action $a^*_i$ is returned by the MPC controller configured with $\parameters$ and sent to the system actuators. This returns a reward $r_i$ that is accumulated in $g_j(\vec x_t)$, where $j$ is the current repetition. Finally, the optimal controller hyper-parameters $\parameters^{*}$ correspond to those with maximum expected cumulative after $n_{BO}$ optimisation iterations. 

\section{Experiments}
\label{sec:experiments}

In this section, we verify the effectiveness of the proposed framework empirically. We evaluate the method, optimising the MPC controller known as MPPI for continuous control problems in both simulated tasks and a physical robot platform. We address two main questions: (Q1) is there a gain over homoscedastic BO (BO$_{homo}$)? (Q2) How does heteroscedastic BO (BO$_{hetero}$) perform against a non-BO baseline that does not take heteroscedasticity into account?

\begin{figure*}[hbt!]
\centering
\input{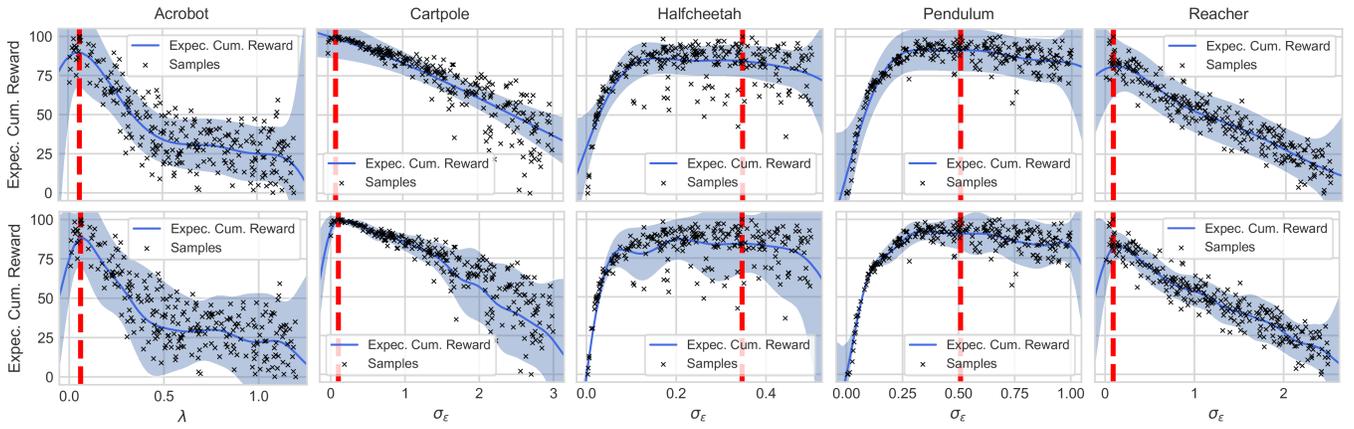}
\caption{Expected cumulative rewards for hyper-parameters sampled via grid search. Homoscedastic GP (upper row) and heteroscedastic GP (lower row). The red dashed line indicates the maximum expected cumulative reward in the sample. The shaded regions correspond to two standard deviations about the mean.
}
\label{fig:rewardcurves}
\end{figure*}

\subsection{Control Problem Simulations}

We conducted experiments on benchmark control problems from OpenAI Gym\footnote{OpenAI Gym: \url{https://gym.openai.com}} \cite{Brockman2016} and Mujoco \cite{Todorov2012}: Acrobot, Cartpole, Half-Cheetah, Pendulum, and Reacher. 

Each control problem has a particular state reward function $r(\vec{\state}, \mathbf{a})$ shown in \autoref{tab:rewards}. We made slight modifications in Reacher and Half-Cheetah. We reduced the effect of actions and gave more priority to the distance to the target in the case of the Reacher problem. For Half-Cheetah, we added more priority to the inclination, since Half-Cheetah would tend to turn upside down as its speed increases. The actuation is then set to finish when such inclination is greater than $\pi/2$ or lower than $-\pi/2$. These modifications make the rewards more informative for MPPI, enabling it to solve these two tasks. We can then focus the analysis on tuning the controller.

The expected cumulative reward represents the expected time the pendulum stays in an upright position in the Acrobot, Cartpole, and Pendulum. It represents the distance traversed in Half-Cheetah, and the speed to reach the target in Reacher. High expected cumulative rewards are the result of motions that increased the reward accordingly, e.g. Half-Cheetah would be expected to reach farther distances.

\begin{table}[t]
\caption{Reward functions used in the experiments. The cartpole and pendulum reward functions were taken from the experiments in \cite{Gardner2017} and the rest from \cite{Wang2019}.}
\label{tab:rewards}
\setlength{\tabcolsep}{0.41em} 
\renewcommand{\arraystretch}{1.2} 
\centering
\begin{tabular}{|c|c|c|}
\hline
\rowcolor{gainsboro}
Control Problem & State Reward \\
\hline
Acrobot & $\cos \state_{1, t} - \cos \left(\state_{1, t}+\state_{2, t}\right)$ \\
\hline
Cartpole & $-(\state_{1,t}^2 + 500 \sin{\state_{3,t}}^2 + \state_{2,t}^2 + \state_{4,t}^2$) \\
\hline
Half-Cheetah & $\dot{s}_{t} - 0.01\left\|\mathbf{a}_{t}\right\|_{2}^{2} - inclination_t$ \\
\hline
Pendulum & $-(50 (\cos \state_{t} - 1)^2 + \dot{\state}^{2}) + 4000$ \\
\hline
Reacher & $-distance_{t} - 0.01 \left\|\mathbf{a}_{t}\right\|_{2}^{2})$ \\
\hline
\end{tabular}
\end{table}

Now, to evaluate the expected cumulative rewards for each problem, we determined fixed values for time horizon $T$, number of trajectory rollouts $M$, and MPPI hyper-parameter intervals are shown in \autoref{tab:search_spaces}. These were found by narrowing down large-enough intervals from near-zero values to 500, taking into account usual values for these hyper-parameters that tend to be close to 0. These are typical in several applications~\cite{Liang2019,Williams2016}. The table also shows optimal values found within these narrowed intervals via grid search.

\newcolumntype{g}{>{\columncolor{gainsboro}}c}
\begin{table}[t]
\caption{MPPI hyper-parameter search spaces and optimal values within the intervals per control problem.}
\label{tab:search_spaces}
\centering
\setlength{\tabcolsep}{0.41em} 
\renewcommand{\arraystretch}{1.15} 
\begin{tabular}{|c|c|c|c|c|c|c|}
\hline
\rowcolor{gainsboro}
Problem & $T$ & $M$ & $\lambda$ interval & $\sigma_\epsilon$ interval & Opt. $\lambda$ & Opt. $\sigma_\epsilon$ \\
\hline
Acrobot      & 8         & 30             & $[10^{-10}, 1.2]$ & $[10^{-10}, 10.0]$ & 0.063 & 8.421  \\
\hline
Cartpole     & 10        & 100            & $[10^{-10}, 1.2]$ & $[10^{-10}, 3.0]$  & 0.757 & 0.158       \\
\hline
Half-Cheetah & 14        & 10             & $[10^{-10}, 0.1]$ & $[10^{-10}, 2.5]$  & 0.026 & 0.263       \\
\hline
Pendulum     & 10        & 10             & $[10^{-10}, 1.2]$ & $[10^{-10}, 3.0]$  & 0.694 & 1.579       \\
\hline
Reacher      & 10        & 15             & $[10^{-10}, 0.1]$ & $[10^{-10}, 2.5]$ & 0.005 & 0.131\\
\hline
\end{tabular}
\end{table}

\subsection{BO Hyper-parameter Search Space and Function Scaling}

For better comparison, both BO variations were implemented using the same squared exponential kernel $k(x, x') = \sigma_n^2\exp\left(-\frac{(x - x')^2}{2\ell^2}\right)$ and UCB acquisition function from \eqref{eq:ucb} with $\kappa=1.2$. For the  optimisation of the acquisition function $h$ we used L-BFGS-B \cite{byrd1995limited} with random starting points as a global optimisation method. We also maximised the marginal likelihood to find the GP hyper-parameters $\theta := \{\sigma,\sigma_n,\ell\}$ for BO$_{homo}$ and $\theta := \{z, \sigma_n,\ell\}$ for BO$_{hetero}$ also using L-BFGS-B. $\theta$ was kept fixed after it was optimised.

Both BO variations were optimised by maximising the GP marginal likelihood of previously observed sample points, which were generated from the defined hyper-parameter intervals from \autoref{tab:search_spaces} for each problem.

\begin{figure}
    \centering
    \subfloat[Degree 1]{\includegraphics[height=2.38cm]{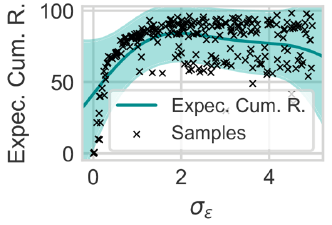}\label{fig:deg1}}
    \subfloat[Degree 5]{\includegraphics[height=2.38cm]{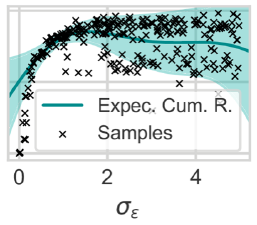}\label{fig:deg5}}
    \subfloat[Degree 10]{\includegraphics[height=2.38cm]{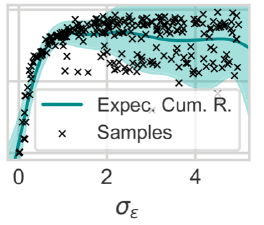}\label{fig:deg10}}
    \caption{Noise model with different polynomial degrees.}
    \label{fig:noises}
\end{figure}

A polynomial basis function $\vec{\phi}$ was used for BO$_{hetero}$ and evaluated for different degrees. A polynomial degree of 1 as in \autoref{fig:deg1} and 5 as in \autoref{fig:deg5} would result in a noise model ignoring small variances while a higher degree would not. We then set a 10-degree polynomial model \autoref{fig:deg10} because it is the first high degree that correctly handles the increasing variance.

The noise model was computed using the regression model in \eqref{eq:reward-approx}. Something else to note about the GP hyper-parameters is that their values have to be proportional to the range of the expected cumulative reward function to model for standard comparison among functions, so the expected cumulative rewards were scaled to $[0, 100]$. 

\subsection{Heteroscedastic Noise Evaluation}

In \autoref{fig:rewardcurves}, we visualise and evaluate the varying noise behaviour of the expected cumulative reward for intervals of MPPI hyper-parameters. We can see that the noise around the mean increases with the $x$-axis hyper-parameter. The expected cumulative reward function for varying temperatures $\lambda$ can be seen only in the Acrobot example, as temperature variations were not significant in the rest of the problems.

The noise heteroscedasticity is evident in all the control problems, so we answer \textbf{Q1}. There is a gain over BO$_{homo}$ as more noise is captured.
However, in Half-Cheetah and Pendulum, the noise is quite skewed around the mean, which means that the expected cumulative reward may not be Gaussian in those cases. The framework still captures a Gaussian noise for the rest of the control problems.

\begin{figure*}[t]
\centering
\includegraphics[height=3.28cm]{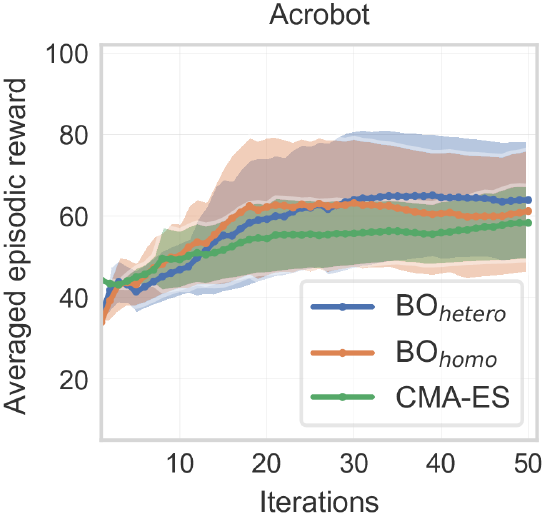}
\includegraphics[height=3.28cm]{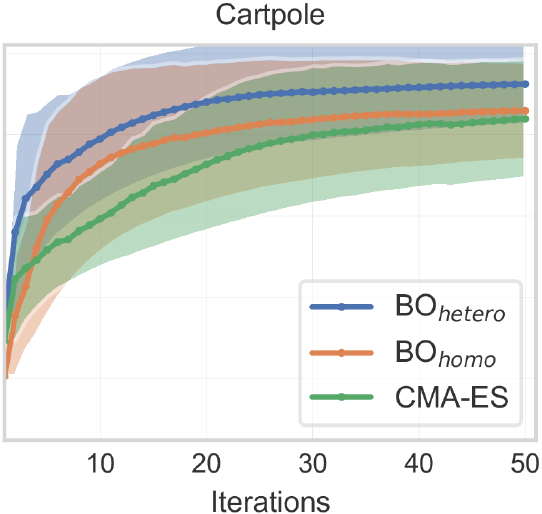}
\includegraphics[height=3.28cm]{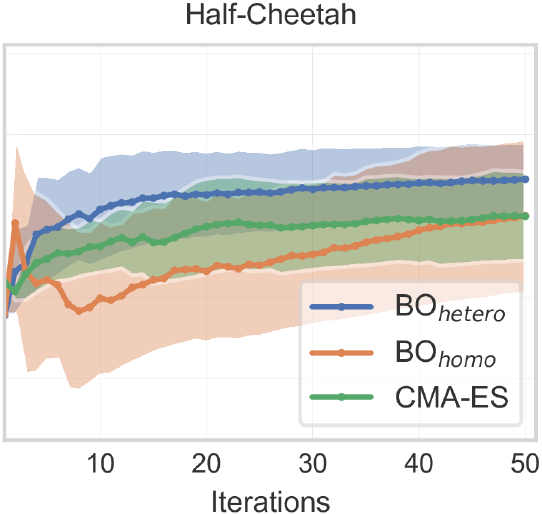}
\includegraphics[height=3.28cm]{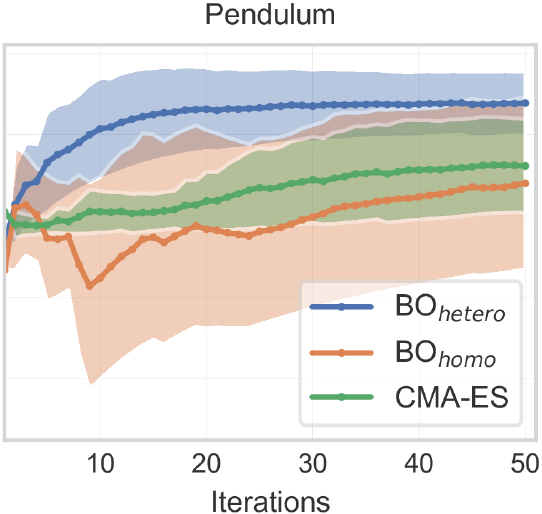}
\includegraphics[height=3.28cm]{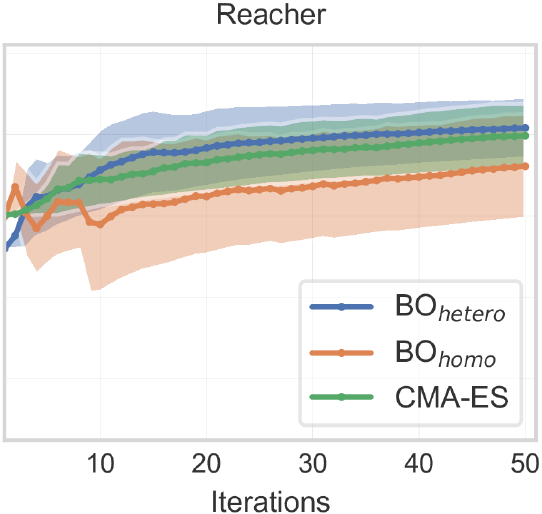}
\caption{Optimisation performance. Results were averaged over 50 episodes with shaded areas and error bars corresponding to two standard deviations. Each method started at the same predefined point in the search space.}
\label{fig:iteration_plots}
\end{figure*}

\subsection{Method Comparison}

To answer \textbf{Q2}, in \autoref{fig:iteration_plots}, we compare BO$_{hetero}$, BO$_{homo}$, and covariance matrix adaptation evolution strategy (CMA-ES) \cite{Arnold2010}, which is a non-BO baseline that does not take heteroscedasticity into account. To allow for proper comparison, each method started at a single defined point in the search space where there's a minimum.
We use CMA-ES with $\sigma_0=1$ and population size of $2$. CMA-ES has been used for hyper-parameter tuning and is considered to be a data-efficient black-box optimiser \cite{loshchilov2016cma, Karro2017}. 

As expected, BO$_{hetero}$ overcomes BO$_{homo}$ and CMA-ES. For BO$_{homo}$, the standard deviation reflects incorrect noise modelled in some regions as also shown in \autoref{fig:rewardcurves}. BO$_{homo}$ ends up with a higher standard deviation in most cases. In Acrobot and Reacher, we did not find much improvement due to mostly homoscedastic regions in the sample collected.

\begin{figure}
    \centering
    \includegraphics[width=.45\textwidth]{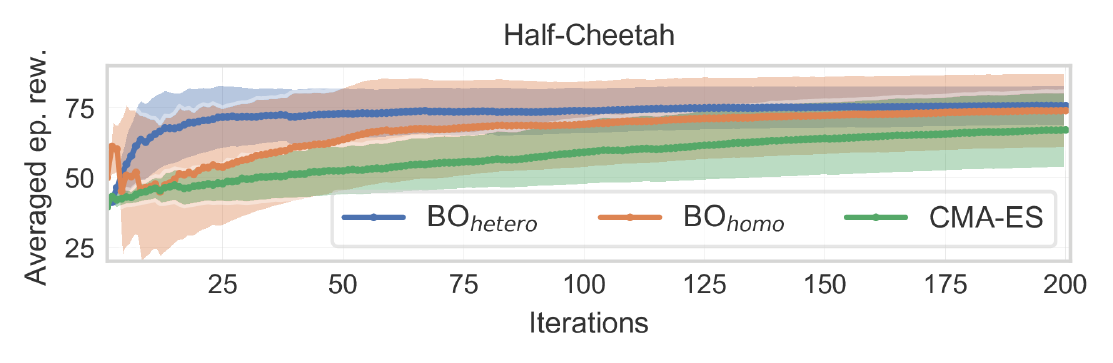}
    \caption{Performance for Half-Cheetah in 200 iterations.}
    \label{fig:longrun}
\end{figure}

To assess long-term performance, we let the optimisation continue for 200 iterations for Half-Cheetah in \autoref{fig:longrun}. As BO$_{hetero}$ describes the overall noise behaviour, it finds optimal regions faster than CMA-ES.

\begin{figure}
    \centering
    \subfloat[Unmodified Half-Cheetah]{\includegraphics[width=.492\columnwidth]{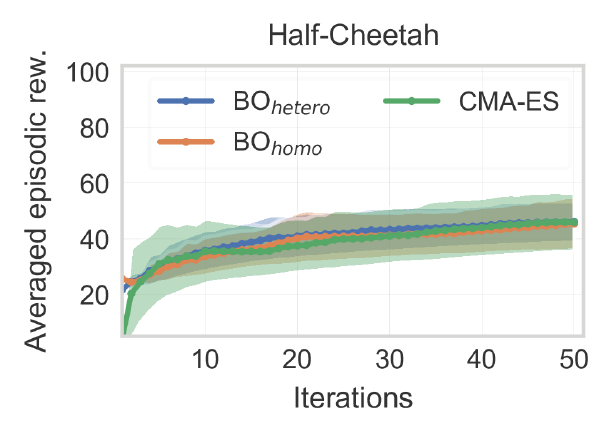}}
    \subfloat[Unmodified Reacher]{\includegraphics[width=.492\columnwidth]{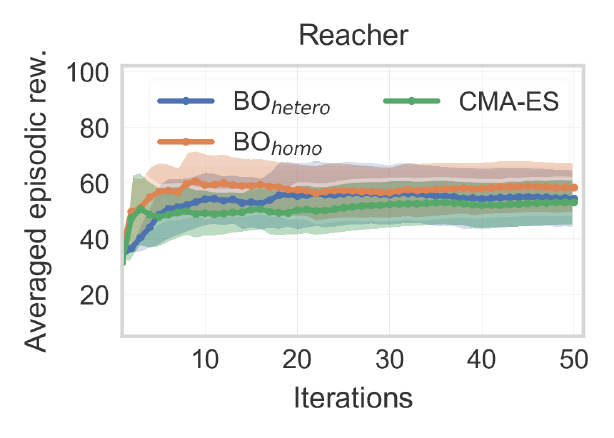}}
    \caption{
    Performance using the unmodified reward functions.}
    \label{fig:unmods}
\end{figure}

It is important to note that CMA-ES does not run inference from prior data. BO is able to apply prior knowledge encoded in the noise model to outperform CMA-ES in fewer iterations. BO approaches do more global search in fewer trials, which is the desired behaviour for a data-efficient solution.

We experimented optimising Half-Cheetah and Reacher with their unmodified reward functions in \autoref{fig:unmods}. The unmodified reward functions make the tasks difficult to all methods, which suggests MPPI has difficulties in solving these tasks. A possible reason is that the unmodified cost function is too uninformative for the MPPI controller.

\subsection{Experiments with a physical robot}

\begin{figure}
    \centering
    \subfloat[Robot]{\includegraphics[width=0.45\columnwidth]{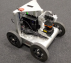}\label{fig:wombot}}
    \subfloat[Noise model]{\includegraphics[width=0.55\columnwidth]{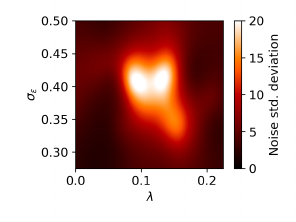}\label{fig:wombot-noise}}

    \subfloat[Results]{\includegraphics[width=0.5\columnwidth]{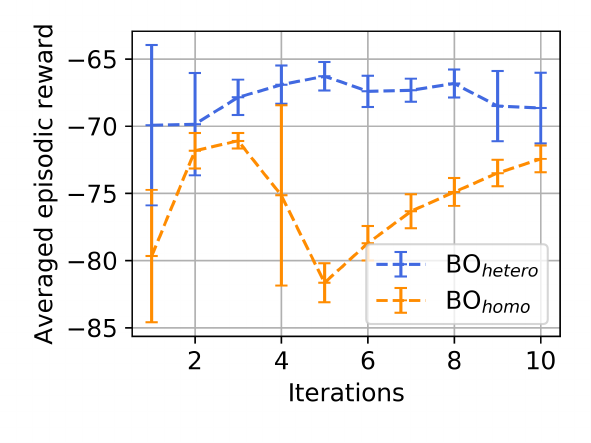}\label{fig:wombot-performance}}
    \caption{Experiments with a physical robot: \protect\subref{fig:wombot} the robot; \protect\subref{fig:wombot-noise} the learnt heteroscedastic noise model; and \protect\subref{fig:wombot-performance} the resulting performance of each BO algorithm. The results were averaged over 3 independent trials for each algorithm, totalling 60 runs of MPPI in 20-second episodes on the robot.}
    \label{fig:robot}
\end{figure}

\begin{figure}
    \centering
    \subfloat[Heteroscedastic GP]{\includegraphics[width=0.903\columnwidth]{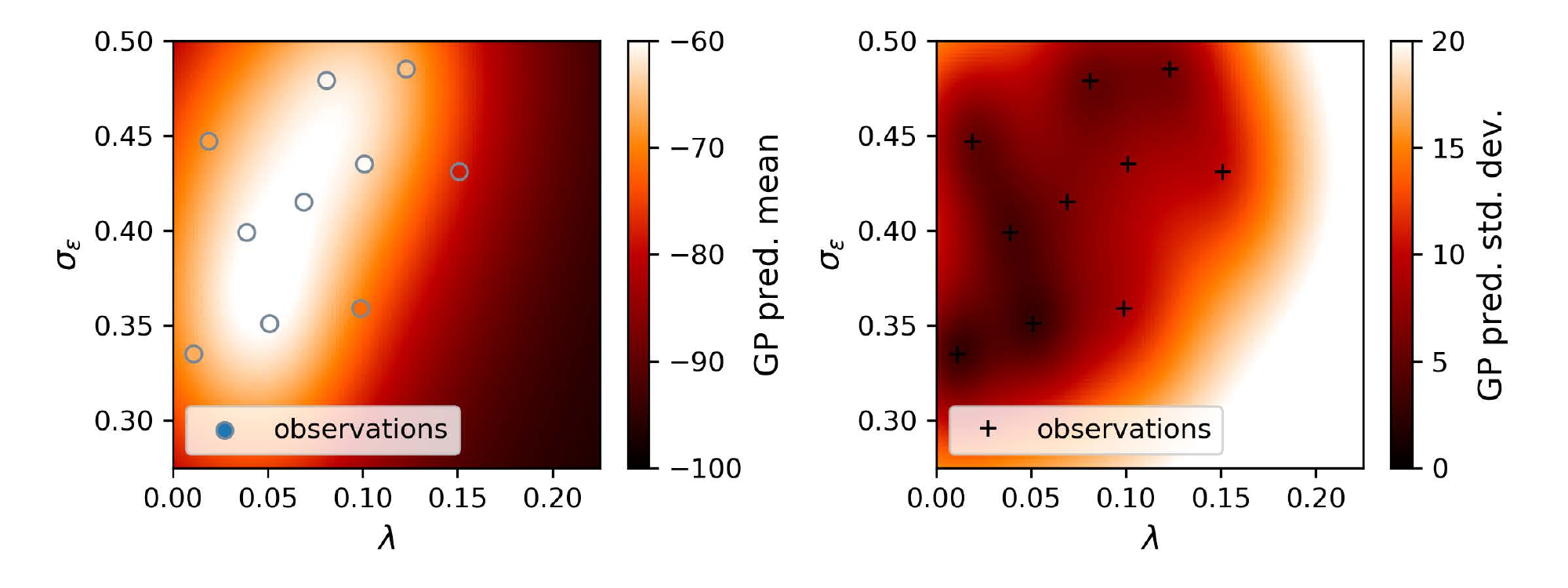}\label{fig:wombot-hom-gp}}

    \subfloat[Homoscedastic GP]{\includegraphics[width=0.903\columnwidth]{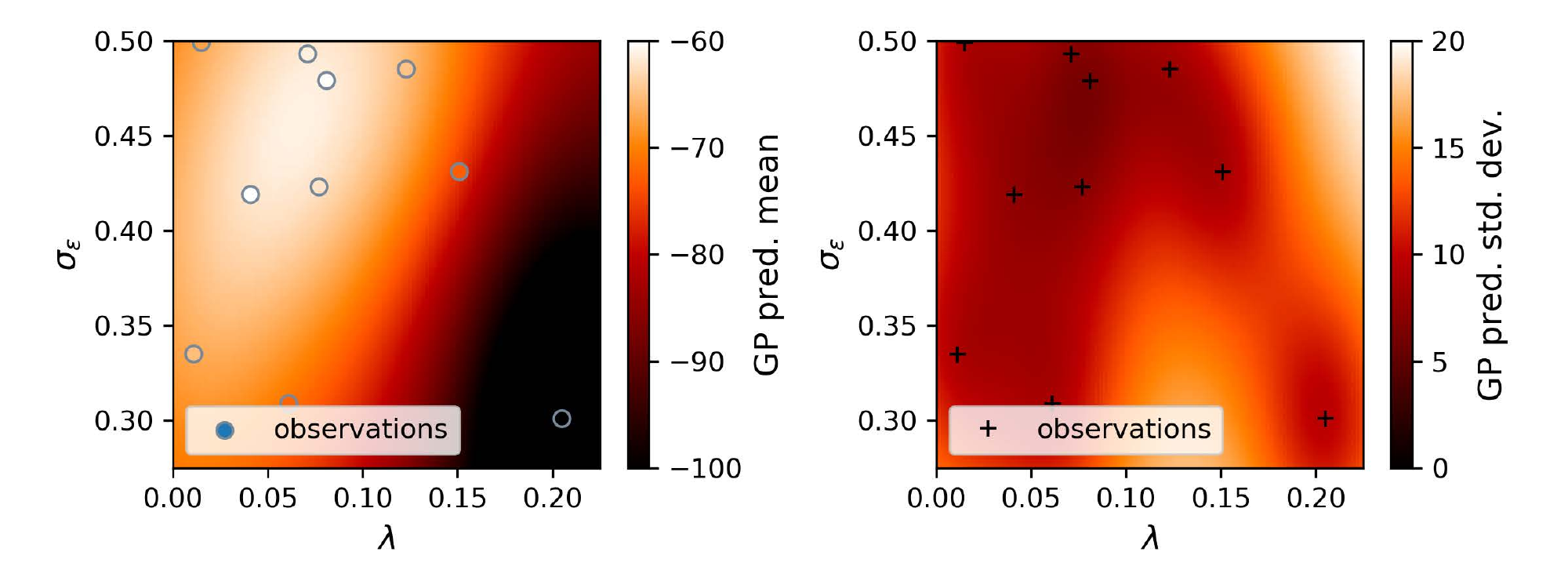}\label{fig:wombot-het-gp}}
\caption{GP models fit with data from one of the trials in the experiments with a physical robot.}
    \label{fig:wombot-gp}
\end{figure}

To assess the effects of real heteroscedastic noise in a physical system, we performed experiments on tuning an MPPI controller for a physical robot. The four-wheel-drive skid-steer robot (\autoref{fig:wombot}) was tasked with following a circular path at a set speed. The cost function was formulated as $\cost(\state_t) = \sqrt{d_t^2+(v_r-v_t)^2}$, where $d_t$ represents the robot's distance to the edge of the circle, $v_r=0.2$ m/s is a reference linear speed, and $v_t$ is the current speed. The robot was localised using a particle filter on a prebuilt map. Internally, MPPI employed a kinematic model of the robot \cite{Kozlowski2004} for trajectory rollouts which is challenging for MPC as the model does not simulate the dynamics of skid-steering platforms accurately. The controller was configured with $M=50$ rollouts and a time horizon $T=400$. Episodes lasted 20 seconds with the robot starting from a fixed initial position. The search space $\Sspace$ for BO was set as the box defined by the intervals $\sigma_\epsilon \in [0.3, 0.5]$ and $\lambda \in [0.01, 0.21]$.

\autoref{fig:wombot-noise} presents the learnt heteroscedastic noise model. Data from preliminary runs revealed that the noise in the episode rewards had a concentrated region of high-variance in roughly the middle of the search space. As previously discussed, both the temperature $\lambda$ and the control noise variance $\sigma_\epsilon^2$ influence MPPI's exploration-exploitation trade-off. For this experiment, the bounds for $\sigma_\epsilon$ were chosen as settings that lead to acceptable performance in practice, but we allowed for a $\lambda$ range which could cause instability. High temperatures $\lambda$ cause excessive exploration in the action space of MPPI, which leads to an almost-sure failure in execution. Conversely, low temperatures force MPPI to take actions that are close to optimal, leading to mostly high rewards. The middle ground between temperature extremes, however, is the region where behaviour is unstable. MPPI's control variance $\sigma_\epsilon^2$ contributes to this behaviour in a similar fashion by determining the spread of the exploration.

To appropriately model the aforementioned noise concentration behaviour, we set the GP noise model as a mixture of stationary kernels by defining (cf. \eqref{eq:noise-model}):
\begin{align}
    \vec\phi(\parameters) &:= [k_q(\parameters, \parameters_1), \dots, k_q(\parameters, \parameters_\nFeatures)]^\transpose~,
\end{align}
where the coefficients $\zeta \in \R$, the weights $\vec\beta\in\R^\nFeatures$ and the points $\parameters_i \in \Sspace$, alongside the other GP hyper-parameters, were tuned offline by maximum a posteriori estimation\footnote{As reasonable choices for the priors, we set log-Gaussian priors for positive GP hyper-parameters and Gaussian priors for the rest.}. As kernel $k_q$, we used the rational quadratic kernel \cite[p. 87]{Rasmussen2006}.

Performance results are in \autoref{fig:wombot-performance}. We compared BO$_{hetero}$ against BO$_{homo}$. Both algorithms are eventually able to find high reward regions. However, due to its uniform noise model, BO$_{homo}$ is led to a more exploratory behaviour, instead of concentrating on promising regions, as evidenced by the query locations in \autoref{fig:wombot-gp}. As a consequence, we observe a significant drop in performance during the optimisation, as shown in \autoref{fig:wombot-performance}. In contrast, BO$_{hetero}$ maintains a steady high performance, which means lower tracking error with respect to the circular path specified by the cost function.

\section{Conclusion}
\label{sec:conclusion}
In this work, we presented a framework for tuning stochastic model predictive control hyper-parameters using Bayesian optimisation with heteroscedastic noise models.
The proposed approach was shown to outperform homoscedastic BO and CMA-ES baselines on classic control tasks in simulated and real environments. A simple and flexible parametric noise model, such as a polynomial, was shown to improve the performance in most of the tasks.
As future work, the online learning of the noise model should allow adapting the model to unforeseen situations. Another point is the skewness of the noise distribution, which could be better modelled as a non-Gaussian distribution. Lastly, we hope this work encourages further analysis of heteroscedasticity in stochastic MPC.

\bibliographystyle{IEEEtran}
\bibliography{rl_bo_references}
\end{document}